\documentclass[letterpaper, 10 pt, conference]{ieeeconf}  

\IEEEoverridecommandlockouts                              

\usepackage{graphicx}
\usepackage{caption}
\usepackage{subfigure}
\usepackage{tabularx}
\usepackage{graphicx} 
\usepackage{times} 
\usepackage{amsmath} 
\usepackage{amssymb}  
\usepackage{microtype}
\usepackage{multirow,color}
\usepackage{algorithm}
\usepackage[noend]{algpseudocode}

\usepackage[dvipsnames]{xcolor}

\newcommand{\cmt}[1]{}

\definecolor{orange}{rgb}{1,0.5,0}

\long\def\ignorethis#1{}

\newcommand{\etal}{{\em{et~al.}\ }}

\newcommand{\vc}[1]{\ensuremath{\mathbf{#1}}}

\newcommand{\argmax}{\operatornamewithlimits{argmax}}

\newcommand{\pctab}{\hspace{0.2in}}

\newcommand{\mytilde}{\raise.17ex\hbox{$\scriptstyle\mathtt{\sim}$}}

\title{\LARGE \bf
Multi-task Learning with Gradient Guided Policy Specialization
}

\author{Wenhao Yu, C. Karen Liu, Greg Turk
\thanks{Wenhao Yu, C. Karen Liu and Greg Turk are with the School of Interactive Computing, Georgia Institute of Technology, Atlanta, Georgia, USA}
\thanks{Wenhao Yu is the corresponding author {\tt\footnotesize(wyu68@gatech.edu)}.}
}

\renewcommand{\baselinestretch}{0.98}
\begin{document}

\maketitle
\thispagestyle{empty}
\pagestyle{empty}

\begin{abstract}
We present a method for efficient learning of control policies for multiple related robotic motor skills. Our approach consists of two stages: joint training and specialization training. During joint training, a single neural network policy is trained to perform multiple tasks. This forces the policy to learn a common representation of the different tasks. Then, during the specialization training stage we selectively split the weights of the policy based on a per-weight metric that measures the disagreement among the multiple tasks. By splitting part of the control policy, it can be further trained to specialize to each task. To update the control policy during learning, we use Proximal Policy Optimization. We evaluate our approach on two continuous control problems in simulation: 1) training three single-legged robots that have considerable difference in shape and size to hop forward and 2) training a 2D biped robot to walk forward and backward. We test our method with different joint training iterations and specialization amounts and compare our method to a random specialization scheme and a standard specialization scheme. Finally, we design a multi-task problem where the inter-task similarity can be continuously controlled. We show that our method can help learning multi-task problems for different types of robots and with different levels of similarities among the tasks.
\end{abstract}


\section{INTRODUCTION}

Deep reinforcement learning (DRL) has achieved considerable success in high dimensional robotic control problems~\cite{LillicrapHPHETS15, schulman2015trust, levine2016end}. Most of these methods have been demonstrated on a single agent that is learning a single task. However, to obtain a truly intelligent agent, it would be desired to train the agent to perform a variety of different tasks, which is referred to as multi-task learning. 

There are three general approaches to train an agent to perform multiple tasks. One can train separate agents for each tasks and later consolidate them together using supervised learning. However, training independently for each task can be sample inefficient. Another approach is to train multiple tasks sequentially. This approach is attractive in that it resembles how humans learn, however it is difficult to design efficient and scalable algorithms that retain the knowledge from earlier tasks. The third category is to learn multiple tasks concurrently. Existing work in this direction has been focused on learning common representations between multiple tasks and trying to achieve better data efficiency.


In this work, we study how weight sharing across multiple neural network policies can be used to improve concurrent learning of multiple tasks. We propose an algorithm that, given a set of learning problems and a fixed data budget, selects the best weights to be shared across the policies. We compute a metric for each weight in the neural network control policy that assesses the disagreement among the tasks using the variance of policy gradients.


We evaluate our methods on learning similar continuous motor control problems using reinforcement learning. We present three examples, and in each example a set of policies are trained on related tasks. We compare our result to four baseline methods, where the first one shares all the weights between policies, the second one shares no weight between policies, the third one randomly selects the weight to split, and the last one uses a standard architecture for learning multi-task problems. We show that by sharing part of the weights selected by our approach, the learning performance can be effectively improved.

\section{RELATED WORK}

In recent years, researchers have used deep reinforcement learning on continuous control problem with high-dimensional state and action spaces~\cite{schulman2015trust, levine2016end, schulman2017proximal}. Powerful learning algorithms have been proposed to develop control policies for highly dynamic motor skills in simulation~\cite{schulman2015trust,schulman2017proximal} or for robot manipulation tasks on real hardware~\cite{levine2016end}. These algorithms typically require a large number of samples to learn a policy for a single task. Directly applying them to train a policy that is capable of multiple tasks might be possible in theory but would be data and computationally inefficient in practice.

One way to train an agent to perform multiple tasks is to first learn each individual task and later consolidate them into one policy. Rusu \etal introduced policy distillation to compress a trained policy into a smaller model or to consolidate multiple trained expert policies into a unified one \cite{rusu2015policy}. They demonstrated that the distilled policy can, in some cases, perform better than the original expert policy. A similar algorithm was proposed by Parisotto \etal to learn a single agent capable of playing multiple Atari games \cite{parisotto2015actor}. Researchers have also applied this approach to learn parameterized robotic control tasks such as throwing darts \cite{da2012learning} or hitting a table tennis ball \cite{kober2011reinforcement}. These algorithms work well when the expert policies are easy to learn individually and do not present conflicting actions, but these assumptions are not always true.

Alternatively, an agent can learn a single policy for multiple tasks sequentially \cite{kirkpatrick2017overcoming, FernandoBBZHRPW17, RusuRDSKKPH16}. Rusu \etal proposed to use a progressive neural network, in which each column corresponds to a task~\cite{RusuRDSKKPH16}. When learning a new task, the algorithm utilizes weights from the previously trained models. Fernando \etal introduced pathnet, which selects pathways from a collection of connected neural network modules for learning new tasks~\cite{FernandoBBZHRPW17}. These methods can effectively retain the knowledge of previously trained policies, but the size of the network can grow extensively. Kirkpatrick \etal proposed to use a quadratic penalty on the neural network weights to prevent the old tasks from being forgotten~\cite{kirkpatrick2017overcoming}. They demonstrated sequential learning results on supervised learning problems and reinforcement learning on Atari games. However, it is unclear how well this method can perform on robotic control problems with a continuous action space.

Directly learning multiple tasks simultaneously has also been well explored \cite{pinto2017learning,yang2017multi,borsa2016learning,teh2017distral}. Pinto and Gupta~\cite{pinto2017learning} demonstrated that simultaneously training two deep neural networks with a partially shared representation achieves better performance than training only one task using the same amount of training data \cite{pinto2017learning}. They manually selected the network parameters to be shared by the two policies, whereas, in our work, we attempt to identify them in an automatic way. For an agent performing multiple tasks in the same environment, Borsa \etal introduced an algorithm to learn the value function with a shared representation and used a multi-task policy iteration algorithm to search for the policy \cite{borsa2016learning}. However, a new value function would need to be trained when a new environment is introduced. Teh \etal applied the idea of policy distillation to multi-task learning by learning a distilled policy that contains common information for all the individual tasks and using it to regularize the learning of task-specific policies\cite{teh2017distral}. They evaluated their method on a maze navigation problem and a 3D game playing problem. However, for controlling robots with potentially different morphologies and joint types, it is unclear whether the common knowledge can be captured in one distilled policy. 

Another line of research for multi-task learning incorporates the task-related information to the state input~\cite{heess2016learning, yu2017preparing, peng2017deeploco, florensa2017stochastic}. Peng \etal duplicated the first layer of the neural network corresponding to different phases of humanoid locomotion~\cite{peng2017deeploco}. A similar architecture was used in~\cite{florensa2017stochastic} to achieve different behaviors in a neural network. Our approach generalizes these architectures by selectively share the weights among the policies.


\section{BACKGROUND}

\subsection{Markov Decision Process (MDP)}

We model robotic control problems as Markov Decision Processes (MDPs), defined by a tuple $(\mathcal{S}, \mathcal{A}, r, \rho_0, P, \gamma)$, where $\mathcal{S}$ is the state space, $\mathcal{A}$ is the action space, $r: \mathcal{S} \times \mathcal{A} \mapsto \mathbb{R}$ is the reward function, $\rho_0$ is the initial state distribution, $P: \mathcal{S} \times \mathcal{A} \mapsto \mathcal{S}$ is the transition function and $\gamma$ is the discount factor. The goal of reinforcement learning is to search for the optimal policy  $\pi_\theta: \mathcal{S} \times \mathcal{A} \mapsto \mathbb{R}$ parameterized by $\theta$, that maximizes the expected long-term reward:
\begin{equation}
\pi_{{\theta}^*} = \argmax_\theta \;\;\mathbb{E}_{\vc{s}{\sim}\rho_0}[V^\pi(\vc{s})],
\label{eq:mdp_obj}
\end{equation}
where the value function of a policy, $V^\pi: \mathcal{S} \mapsto \mathbb{R}$, is defined as the expected long-term reward of following the policy $\pi_\theta$ from some input state $\vc{s}_t$: 
\begin{equation} 
V^\pi(\vc{s}_t) = \mathbb{E}_{\vc{a}_t \sim \pi {,}\vc{s}_{t+1} \sim P{,}...}[\sum_{i=0}^\infty \gamma^i r(\vc{s}_{t+i}, \vc{a}_{t+i})].
\end{equation}

In the context of multi-task learning, we can model each task as an MDP $(\mathcal{S}_i, \mathcal{A}_i, r_i, \rho_{0_i}, P_i, \gamma_i)$, where $i$ is the index of the $i^{th}$ task. The objective function (\ref{eq:mdp_obj}) is then modified to be:

\begin{equation}
\pi_{{\theta}^*} = \argmax_\theta \;\;\frac{1}{N}\sum_i^N\mathbb{E}_{\vc{s}{\sim}\rho_{0_i}}[V^\pi_i(\vc{s})],
\end{equation}
where $N$ is the total number of tasks.

\subsection{Policy Gradient Algorithm}

Policy gradient methods directly estimate the gradient of the objective function (\ref{eq:mdp_obj}) w.r.t the policy parameters $\theta$ and use gradient descent to optimize the policy. In this work, 
we use Proximal Policy Optimization (PPO) \cite{schulman2017proximal} to learn the optimal control policy because it provides better data efficiency and learning performance than the alternative learning algorithms.

Similar to many policy gradient methods, PPO defines an advantage function as $A^\pi(\vc{s}_t, \vc{a}) = Q^\pi(\vc{s}, \vc{a}) - V^\pi(\vc{s})$, where $Q^\pi$ is the state-action value function that evaluates the return of taking action $\vc{a}$ at state $\vc{s}$ and following the policy $\pi_\theta$ thereafter. However, PPO minimizes a modified objective function to the original MDP problem:
\begin{align}
L_{PPO}({\theta}) = -\mathbb{E}_{\mathcal{R}} [\min (&r(\theta)A_t, clip(r(\theta), 1-\epsilon, 1+\epsilon)A_t)],
\label{eq:ppo_loss}
\end{align}
where $\mathcal{R}=[\vc{s}_0, \vc{a}_0, ...]$ are rollouts collected using an old policy $\pi_{\theta_{old}}$, $r(\theta) = \frac{\pi_\theta(\vc{a}|\vc{s})}{\pi_{\theta_{old}}(\vc{a}|\vc{s})}$ is the importance re-sampling term that enables us to use data sampled under the old policy to estimate expectation for the current policy $\pi_{\theta}$. The $min$ and the $clip$ operators together ensure that $\pi_{\theta}$ does not change too much from $\pi_{\theta_{old}}$. More details in deriving the objective functions can be found in the original paper on PPO \cite{schulman2017proximal}.

\section{METHOD}
Our method aims to integrate two important techniques in learning multiple tasks, joint learning and specialization, into one coherent algorithm. Learning multiple tasks jointly is a well known strategy to improve learning efficiency and generalization. However, to further improve each individual task, a specialized curriculum and training are often needed. We define a ``task'' as a particular reward function performed by a particular dynamic system. Therefore, two different tasks can mean two different robots achieving the same goal, the same robot achieving two different goals, or both.

Our algorithm carries out two learning phases: joint training and specialization training. We first train a policy, $\pi_\theta(\vc{a} | \vc{s}): (\mathcal{S}, \mathcal{A}) \mapsto [0,1]$, represented by a fully connected neural network to jointly learn the common representations across the different tasks (Section \ref{ssec:policy_pretrain}). A policy is defined as a Gaussian probability distribution of action $\vc{a} \in \mathcal{A}$ conditioned on a state $\vc{s} \in \mathcal{S}$. The mean of the distribution is represented by the neural network and the covariance is defined as part of the policy parameters, $\theta$, which also include the weights and the biases of the network.  Based on the gradient information gathering during the  joint training phase, we then select a subset of weights to be specialized to individual tasks (Section \ref{ssec:weight_share}).

\subsection{Joint Training}
\label{ssec:policy_pretrain}

The goal of jointly learning multiple tasks is to learn a common representation of the multiple tasks, as well as to provide critical information to determine which weights should be shared across tasks in the specialization training phase. The training process is identical to training a single task except that the rollout pool consists of trajectories generated for performing different tasks. We use PPO to search for a policy that maximizes the surrogate loss defined in Equation (\ref{eq:ppo_loss}). During joint training, the policy does not distinguish between different tasks. This forces the policy to learn a common representation of the multiple tasks. 

\subsection{Specialization Training}
\label{ssec:weight_share}
In the specialization training phase, we first analyze the policy after the joint training phase and select a subset of the weights to be shared across the policies. We compute a per-weight specialization metric to estimate whether a particular weight in the neural network should be shared or specialized to each task. The key idea of our approach is, for each weight in the network, to estimate the disagreement among different tasks.

Algorithm \ref{alg:weight_share} begins by collecting rollouts $\mathcal{R}_i$ for each task $i$ using the current policy. We then approximate the gradient of the PPO loss (Equation \ref{eq:ppo_loss}) with respect to the policy parameters $\theta$, using rollouts $\mathcal{R}_i$:

\begin{equation}
\vc{g}_i \equiv \frac{\partial L_{PPO}}{\partial \theta}.
\label{eq:ppo_grad}
\end{equation}

After we approximate the policy gradients for all tasks, we obtain $N$ gradient vectors: $\vc{g}_1 \in \mathbb{R}^{|\theta|}, \cdots, \vc{g}_N\in \mathbb{R}^{|\theta|}$. For each element of the gradient vector, we compute its variance across all $N$ tasks. If a particular variance is low, this implies that the tasks are in agreement on the update of the corresponding weight in the network. From $| \theta |$ variances, we identify the $M$ smallest ones and share the corresponding $M$ weights across the policies of different tasks. The rest of the weights will be split to specialize each task. 

\begin{figure}[t!]
\centering
\subfigure{\includegraphics[width=7cm]{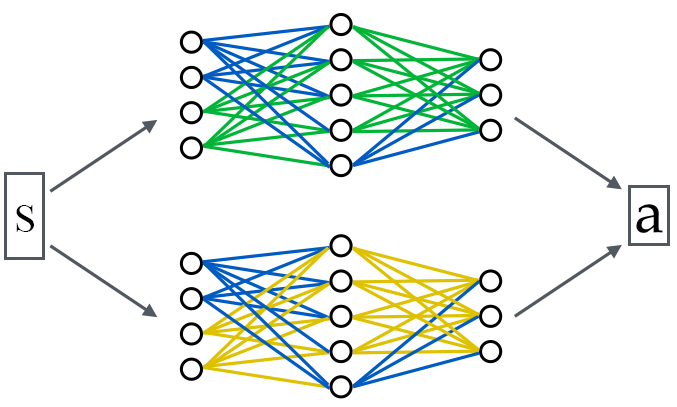}}
\caption{Illustration of network architecture. Blue edges denote the weights shared by the tasks. Green edges are weights specialized to task 1 and yellow edges are specialized to task 2. }
\label{fig:split_illus}
\vspace{-3mm}
\end{figure}

The architecture of the new network can be viewed as $N$ copies of old networks jointed at the output layer to produce the final action. If a weight is shared, its value must be the same across $N$ subnetworks and is updated in unison. Otherwise, a weight can assume different values for the different subnetworks (Figure \ref{fig:split_illus}). We initialize the new network using the values of the old network after the joint training phase. 

Since the policy gradient itself is approximated with samples, the specialization metric depending on these gradients might also contain a significant amount of noise. In our experiments, we average the specialization metric over $10$ iterations of PPO updates before selecting the weights to be shared or specialized.

\begin{algorithm}[t]
\caption{Gradient-guided weight selection}\label{alg:weight_share}
\begin{algorithmic}[1]
\For{$i=1:N$}
\State Collect rollouts $\mathcal{R}_i$ from task $i$\;
\State Compute $\mathbf{g}_i = \frac{\partial L_{PPO}}{\partial \theta}$ using $\mathcal{R}_i$\;
\EndFor
\For{$j=1:|\theta|$}
\State $v_j = $ variance($\vc{g}_1[j], \cdots \vc{g}_N[j]$)\;
\EndFor
\State Select the smallest $M$ $v$'s \;
\State Share corresponding $M$ weights in network\;
\end{algorithmic}
\end{algorithm}

\section{RESULTS}

We evaluate our approach on three multi-task continuous control problems. Our algorithm introduces two hyper-parameters that need to be determined: the number of iterations of joint training (jt) and the percentage of specialized network weights (sp). For each task, we run our method with joint training (jt) of $80k$, $400k$ and $800k$ samples and specialization percentages (sp) of $5\%$, $25\%$ and $50\%$, creating $9$ sets of hyper-parameters in total. We also test our method for joint training only (sp$=0\%$) and $100\%$ specialization without joint training (jt$=0$, sp$=100\%$). We compare our method to two other baseline methods where 1) specialization is done randomly and 2) a one hot vector is appended to the policy input to minimally disambiguate the tasks (append). Among the hyper-parameters being tested, we find that jt$=400k$ and sp$=25\%$ works consistently well for all of our examples. Thus we use these parameters for the random specialization baseline.

We use the implementation of PPO in OpenAI Baselines \cite{baselines}. To represent the control policy, we use a neural network with three hidden layers, comprised of $64$ hidden units each. All results demonstrated in this work are simulated using DartEnv \cite{dartenv}, a fork of the OpenAI Gym \cite{BrockmanCPSSTZ16} library that uses Dart \cite{DART} as the underlying rigid body simulator. The simulation
timestep is set to $0.002$s. We run each example three times and report the average learning curves. We choose the total iteration numbers empirically so that the policies can be sufficiently trained to learn the motor skills.

\subsection{Robot hopping with different shapes}

\begin{figure}[t!]
\centering
\subfigure{\includegraphics[width=7cm]{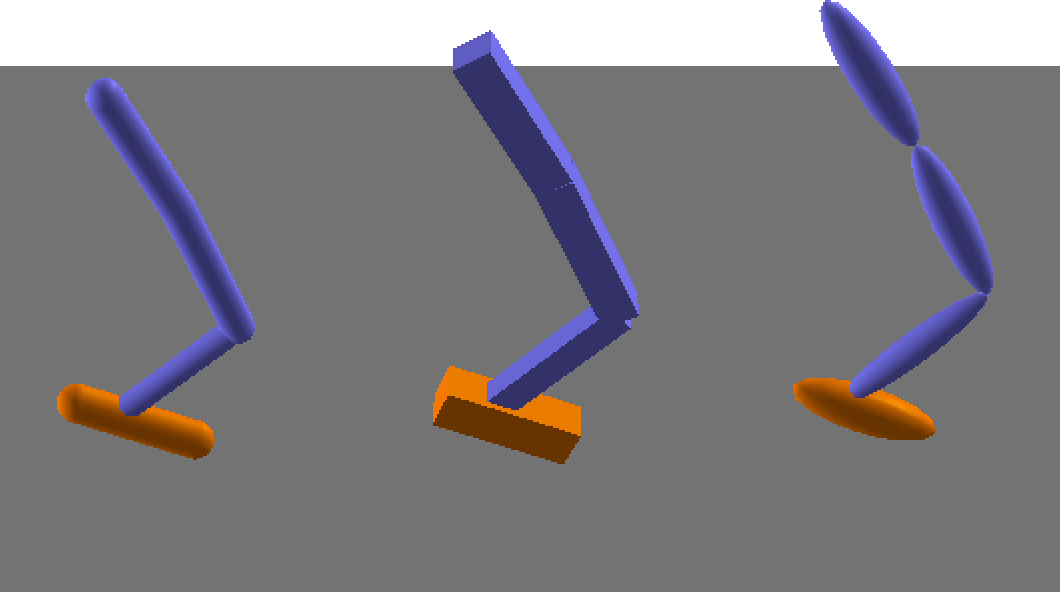}}
\caption{Three single-legged robots with different shapes and sizes. All of them are trained to hop forward.}
\label{fig:3hopper}
\vspace{-3mm}
\end{figure}

We begin with an example of hopping locomotion of a single-legged robot. We design three single-legged robots that are constructed with capsules, boxes and ellipsoids respectively, as shown in Figure~\ref{fig:3hopper}. In addition, we scale them to have different total heights. These variations lead to considerable difference in the inertia and contacts of the robots, while the similarity in the configuration should lead to similar locomotion gaits, which we expect the joint-training to capture. We use a batch size of $4,000$ for the training.

\begin{figure}[t!]
\centering
\subfigure{\includegraphics[width=8.5cm]{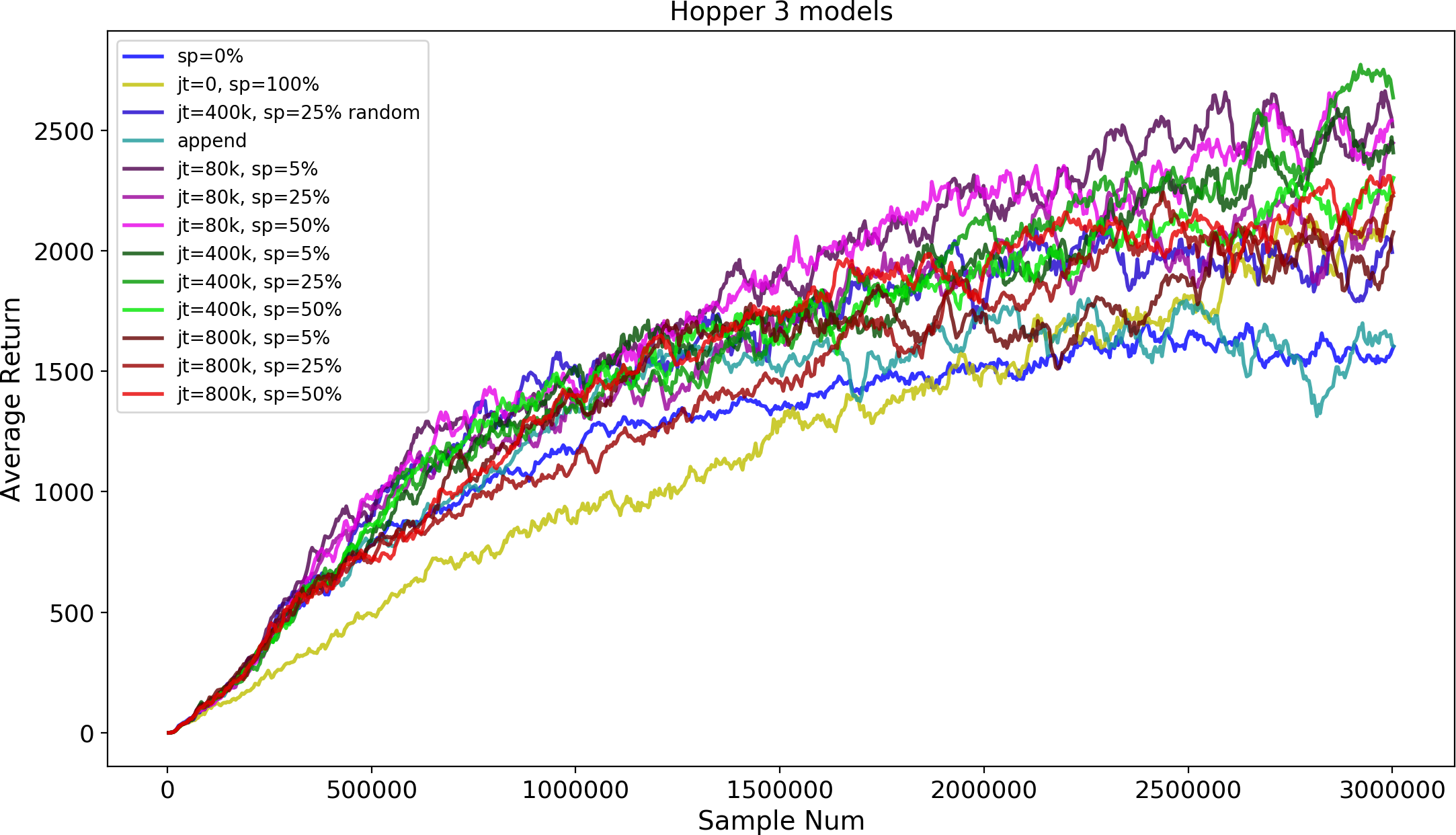}}
\caption{Learning curves for the three hoppers example.}
\label{fig:3hopper_result}
\vspace{-3mm}
\end{figure}

The result of this example can be found in Figure \ref{fig:3hopper_result} and Table \ref{tbl:policy_performance}. In most cases, using joint training with specialization achieves better performance than all four baselines (sp$=0\%$, jt$=0 $ sp$=100\%$, jt$=400k $ sp=$25\%$ random and append), showing the effectiveness of our method for this problem. Meanwhile, we observe that the three learning curves associated with jt$=400k$ obtained the best overall performance, while with jt=$80k$ notable variance can be observed across  different specialization amounts.

\subsection{2D bipedal walking in two directions}

\begin{figure}[t!]
\centering
\subfigure{\includegraphics[width=8.5cm]{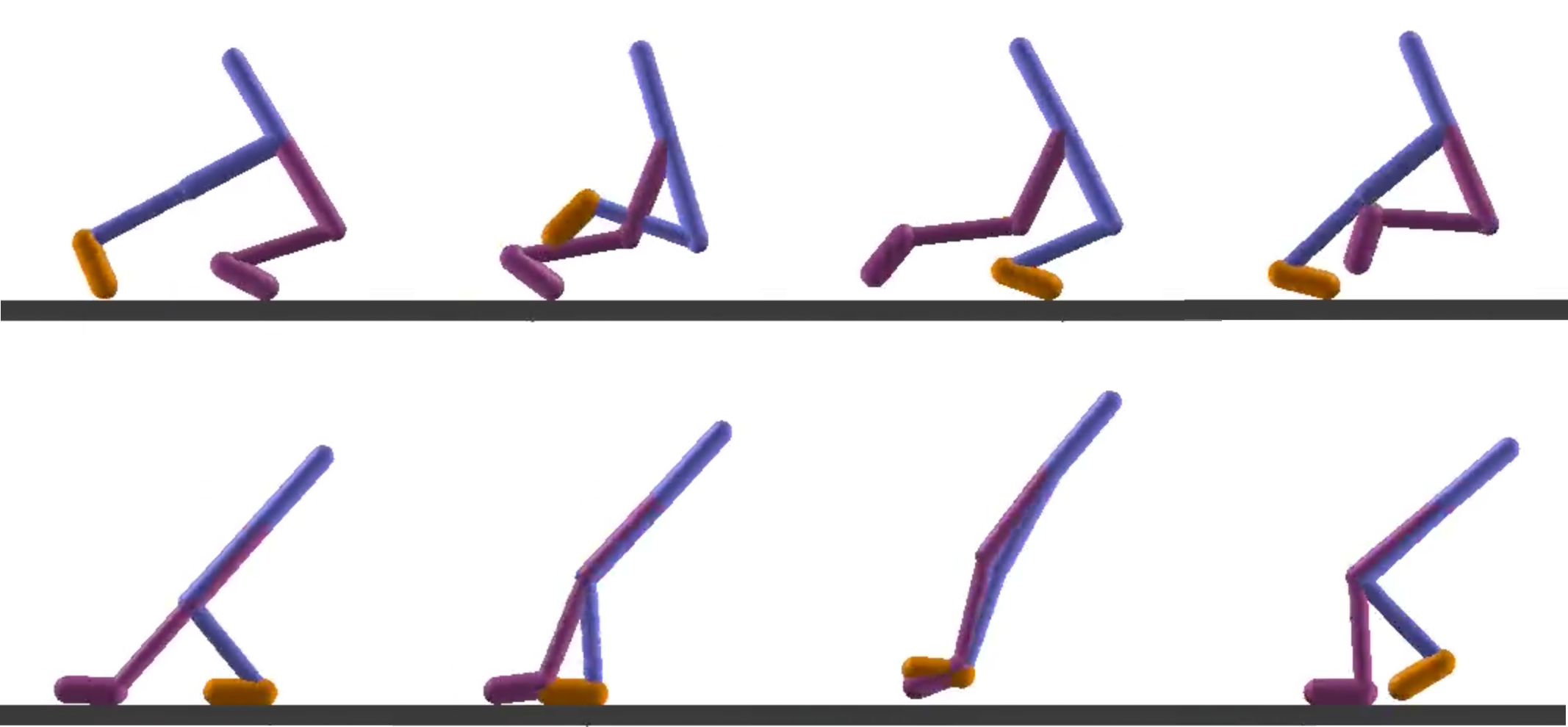}}
\caption{Bipedal robot moving forward (top) and backward (bottom).}
\label{fig:walker2d_illus}
\vspace{-3mm}
\end{figure}

In this example, we train one robot to perform different tasks. Specifically, we train a bipedal robot to move forward and backward. This is different from the previous examples where we train different robots to perform the same task. The bipedal robot is constructed similar to the 2D Walker example in OpenAI Gym \cite{BrockmanCPSSTZ16} and is constrained to move in its sagittal plane. We reward linear velocities at the COM of the robot for being positive and negative to achieve walking forward and backward. We use a batch size of $8,000$ for training the policies. An illustration of the resulting motion can be seen in Figure \ref{fig:walker2d_illus}. 

\begin{figure}[t!]
\centering
\subfigure{\includegraphics[width=8.5cm]{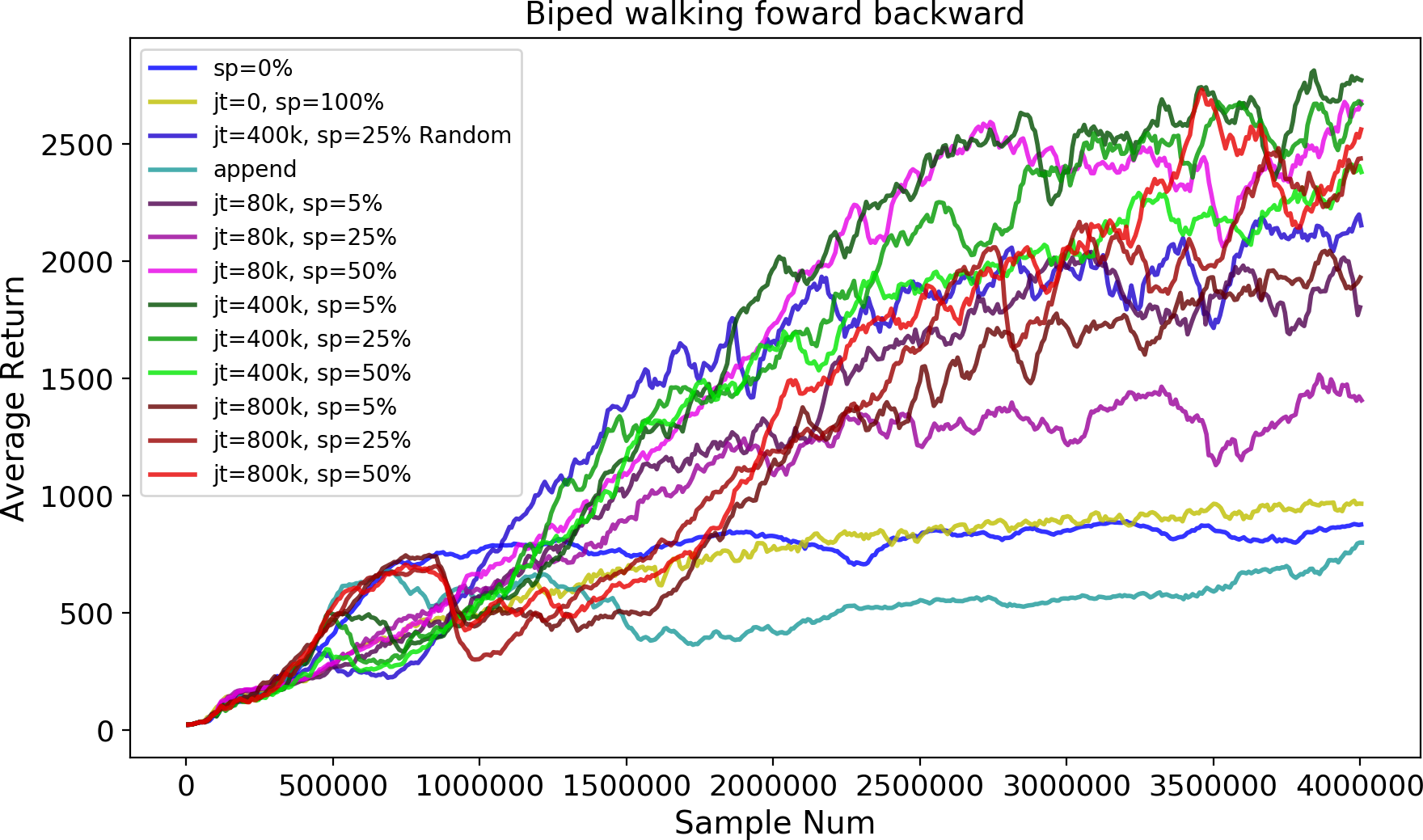}}
\caption{Learning curves for the bipedal robot example.}
\label{fig:walker2d_result}
\vspace{-3mm}
\end{figure}

The learning result of this example is shown in Figure~\ref{fig:walker2d_result} and Table \ref{tbl:policy_performance}. We can see that specialization works particularly well for this task and even random specialization achieves decent learning performance. Note that with the same amount of joint training and specialization, our approach still outperforms random specialization. In addition, though joint training only (sp$=0\%$) and training separately (jt$=0$, sp$=100\%$) achieve similar performance, they learn different behaviors, with the former learn to stand still and the latter learn to make a few steps before losing balance.

\begin{table}[b]
\begin{center}
    \caption{Average final performances of different methods for training the three hopping robots with different shapes (3 Hoppers) and the biped robot walking forward and backward (Biped).}
\label{tbl:policy_performance}
\begin{tabular}{|l|c|c|}
\hline
& 3 Hoppers & Biped \\
\hline
sp$=0\%$ & 1601.94 &877.80\\
\hline
jt=$0$,sp=$100\%$ & 2240.07 &966.65\\
\hline
jt=$400$k,sp=$25\%$ random & 1994.28 & 2154.09\\
\hline
append & 1603.91 & 800.59\\
\hline
 jt=$80$k,sp=$5\%$ & 2516.05 & 1804.13\\
\hline
 jt=$80$k,sp=$25\%$ & 2447.45 & 1407.69\\
\hline
 jt=$80$k,sp=$50\%$ &  2541.31 & 2681.31\\
\hline
 jt=$400$k,sp=$5\%$ & 2407.68 & 2773.29\\
\hline
 jt=$400$k,sp=$25\%$ & 2635.56 & 2671.23\\
\hline
 jt=$400$k,sp=$50\%$ & 2302.74 & 2381.18\\
\hline
 jt=$800$k,sp=$5\%$ & 2076.98 & 1931.47\\
\hline
 jt=$800$k,sp=$25\%$ & 2227.95 & 2438.77\\
\hline
 jt=$800$k,sp=$50\%$ & 2242.20 & 2562.53\\
\hline

\end{tabular}
\end{center}
\end{table}

\begin{figure*}[t!]
\centering
\subfigure{\includegraphics[width=18cm]{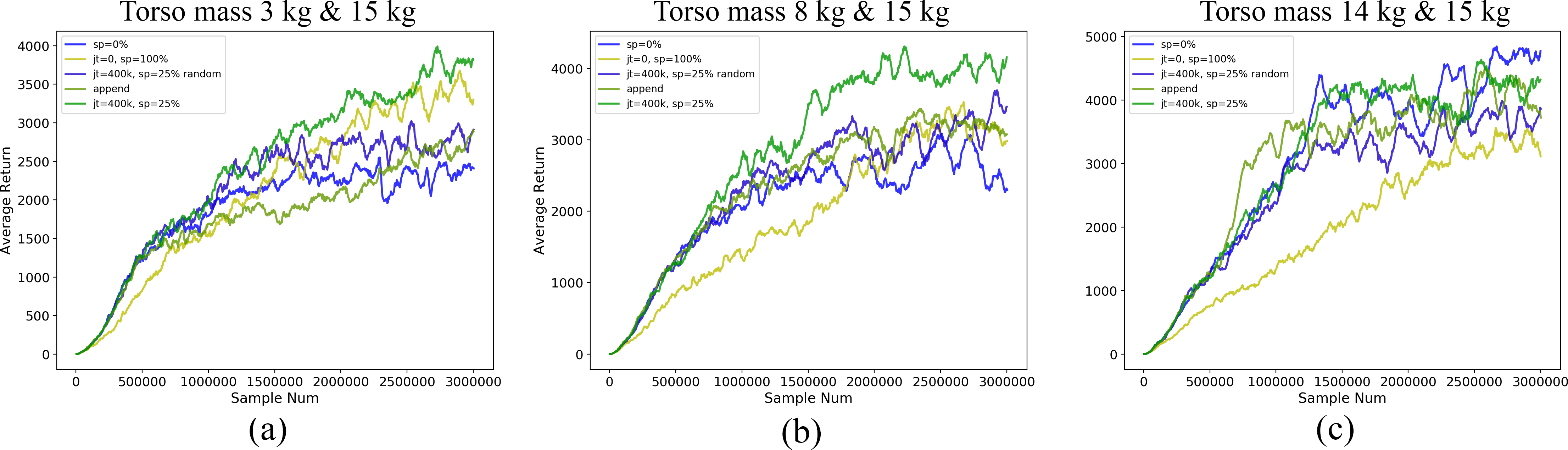}}
\caption{Learning curves for the hopper example. (a) the two hoppers have torso mass of $3$ kg and $15$ kg. (b) the two hoppers have torso mass of $8$ kg and $15$ kg. (c) the two hoppers have torso mass of $14$ kg and $15$ kg.}
\label{fig:torso_mass_result}
\vspace{-3mm}
\end{figure*}

\subsection{Hopper with different torso mass}

In this example, we train two single-legged robot with different torso mass to hop forward. By controlling the difference between the torso mass of the two robots, we can specify the similarity among the tasks for a multi-task problem. We fix one of the two robot to have a torso mass of $15$ kg and assign the torso mass of the other robot from three options: $3$ kg, $8$ kg and $14$ kg. We use a batch size of $4,000$ for training the policies.

The results are shown in Figure \ref{fig:torso_mass_result}. We see that when the two tasks are very different from each other, training two policies separately can achieve better performance than training single policy (Figure \ref{fig:torso_mass_result}(a)). As the two tasks become more similar, training single policy achieves better performance and eventually outperforms training separate policies (Figure \ref{fig:torso_mass_result}(b) and (c)). In any case, the policies trained with our method always achieve top performance in all three problems, showing the effectiveness of joint training and selective specialization.

\section{DISCUSSION}

We have shown that by combining joint training and policy specialization, we can improve the performance and efficiency of concurrent learning of multiple robotic motor skills of different types. We evaluated our method on $9$ sets of hyper-parameters for joint training and specialization.  We demonstrated that in most cases our approach can be helpful compared to baselines, while different joint training and specialization amounts can result in notably different performances. We identified one pair of hyper-parameter that works well for all the presented examples, which in a practical setting, can be used as an initial guess followed by additional fine-tuning of the hyper-parameters. However, we recognize a few limitations that require further investigations. 

In this work, we investigated training with  specialization occurring at one particular point during learning, and do not allow parameters to be shared by a subset of the tasks during specialization. We found this scheme work well for our test cases, however, a potentially more powerful strategy would be to allow multiple specializations throughout the learning and perform specialization at subsequently finer levels.

In PPO, both the policy $\pi$ and the value function $V^{\pi}$ are represented by neural networks and optimized throughout the training. In this work, we apply specialization only to the policy network. We found that specialization of the value function network did not achieve notable improvement in a preliminary test and it would lead to additional hyper-parameter search. However, for certain problems, it could be beneficial to specialize the value function network as well. Our framework can be easily applied to achieve this by replacing the PPO surrogate loss in Equation (\ref{eq:ppo_grad}) with the loss of the value function network.

One important direction of investigation is to automatically determine the optimal hyper-parameters. One possible direction would be to learn a predictive model that estimates the performance improvement for different amounts of specialization. Another future direction is generalization to new tasks. Through training multiple tasks with shared parameters, it is possible that the trained policies learn a common representation of the space of the multiple tasks. It would be interesting to see if training on a novel but related task by initializing the policy using our approach would achieve better learning performance.

\section{CONCLUSION}

We have introduced a method for learning multiple related robotic motor skills concurrently with improved data efficiency. The key stages of our approach consist of a joint training phase and a specialization phase. We proposed a metric using the variance of the task-based policy gradient to selectively split the neural network policy for specialization. We demonstrated our approach on three multi-task examples where different robots are trained to perform different tasks.  For these examples, our approach improves the learning performance compared to joint training alone, independent training, random policy specialization and a standard architecture for multi-task learning.

\addtolength{\textheight}{-12cm}   




\section *{ACKNOWLEDGMENT}

This work is supported by NSF award IIS-1514258.

\bibliographystyle{IEEEtran}
\bibliography{reference_bib}

\end{document}